\documentclass{article}

\PassOptionsToPackage{numbers, compress}{natbib}
\usepackage[preprint]{neurips_2026}

\usepackage[utf8]{inputenc} 
\usepackage[T1]{fontenc}    
\usepackage{hyperref}       
\usepackage{url}            
\usepackage{booktabs}       
\usepackage{amsfonts}       
\usepackage{nicefrac}       
\usepackage{microtype}      
\usepackage{xcolor}         
\usepackage[table]{xcolor}

\usepackage{amsmath}
\usepackage{graphicx}
\usepackage{multirow}
\usepackage{enumitem}
\definecolor{tabblue}{rgb}{0.12,0.49,0.85}

\title{PAR3D: A Unified 3D-MLLM with Part-Aware Representation for Scene Understanding}

\author{%
  Shaohui Dai\thanks{Equal contribution.}~~~
  Yansong Qu\footnotemark[1]~~\thanks{Project Lead.}~~~
  You Shen~~
  Shengchuan Zhang~~
  Liujuan Cao\thanks{Corresponding Author}\\
  Key Laboratory of Multimedia Trusted Perception and Efficient Computing,\\Ministry of Education of China, Xiamen University
}

\begin{document}
\maketitle

\begin{figure}[h]
    \centering
    \includegraphics[width=\textwidth]{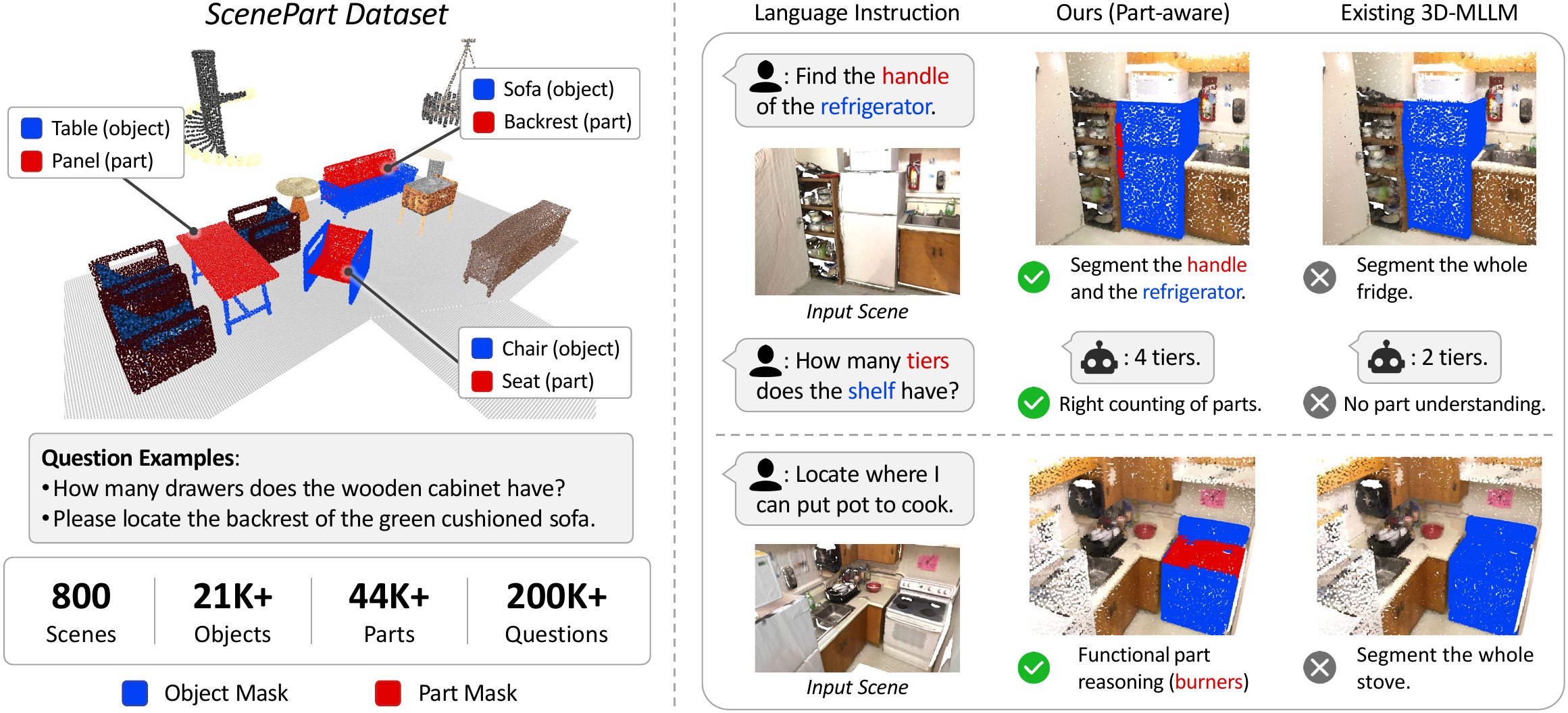}
    \caption{
We propose PAR3D, a unified 3D-MLLM with part-aware representation, together with ScenePart dataset.
\textit{Left:} ScenePart provides fine-grained object-part annotations and language instructions for 3D scenes.
\textit{Right:} PAR3D enables part-aware understanding across question answering, segmentation, and reasoning, going beyond the object-level understanding of existing 3D-MLLMs.
    }
    \label{fig:teaser}
\end{figure}

\begin{abstract}
    Recent advances in 3D multimodal large language models (3D-MLLMs) have enabled unified solutions for 3D scene understanding tasks, including visual question answering, captioning, and referring segmentation. However, existing 3D-MLLMs remain largely object-centric, limiting their ability to model fine-grained part structures that are essential for embodied interaction with 3D environments.
    In this work, we present \textit{PAR3D}, a unified part-aware 3D-MLLM framework that enables models to understand, reason about, and ground both objects and their parts in 3D scenes. 
    To enable training and evaluation of part-aware 3D scene understanding, we introduce ScenePart, a synthetic 3D scene dataset with part-level annotations and language instructions. 
    We further develop Part-Aware 3D Representation Learning to enrich 3D visual representations with fine-grained part-level semantics, 
    and propose Hierarchical Segmentation Query Generation to ground part targets via hierarchical object--part queries.
    Extensive experiments show that our method substantially improves part-level question answering and referring segmentation, while also achieving strong performance across object-level vision-language tasks. Project page: \url{https://atrovast.github.io/PAR3D/}.
\end{abstract}

\section{Introduction}
Language-guided understanding of 3D environments is a fundamental problem in computer vision and a key capability for embodied intelligence. Recent 3D multimodal large language models (3D-MLLMs) have made substantial progress toward this goal by connecting 3D perception modules with large language models, enabling a unified interface for 3D scene perception and spatial reasoning~\cite{ma2024survey1, zha2025survey2, huang2023leo, huang2024chatscene}. These advances make 3D-MLLMs promising for robotics, augmented reality, and digital twins, where intelligent systems must interpret complex scenes and respond to language instructions.

However, many real-world interactions require understanding a scene beyond object-level recognition. An embodied agent may need to grasp the \textit{handle} of a mug, pull the \textit{drawer} of a cabinet, sit on the \textit{seat} of a chair, or inspect the \textit{door} of a refrigerator, requiring affordance-aware perception and manipulation of object parts~\cite{mo2021where2act,geng2023gapartnet}.
Beyond embodied intelligence, similar fine-grained structural awareness is also useful for controllable 3D content editing and interactive scene manipulation, where systems may need to select or modify local functional regions rather than entire objects~\cite{qu2025drag, wang2025partxmllm}.
In these cases, the target of reasoning is not merely an object instance, but a functional component embedded within an object and situated in a scene.
This requires part-aware 3D scene understanding: the model should recognize object parts as meaningful semantic and functional units, understand their dependence on host objects and scene context, and localize target parts within host objects.

Part-aware 3D scene understanding exposes a structural mismatch in existing 3D-MLLMs. Most current approaches are built around an object-centric view of 3D scenes, where objects serve as the primary units for visual representation~\cite{huang2024chatscene, chen2024grounded3dllm, zemskova20253dgraphllm}, language alignment~\cite{hong20233dllm, huang2023leo, fu2024scenellm}, and task supervision~\cite{deng20253dllava, he2024segpoint, huang2025reason3d}. 
A straightforward solution is to extend the object-centric formulation to object parts as finer-grained targets.
Object parts, however, are not simply smaller objects. They are structured components whose meanings are tied to their host objects and whose functions often determine how an agent should interact with them. 
As illustrated in Figure~\ref{fig:teaser}, existing 3D-MLLMs may identify the host object but overlook its functional parts. 
This object-centric bias manifests in three key aspects. 
First, existing 3D datasets~\cite{dai2017scannet, Wald20193rscan, baruch2021arkitscenes} largely lack part-level annotations in scene context, making it difficult to supervise object--part understanding. 
Second, existing 3D visual backbones are typically adapted from object-level scene perception and may not preserve the fine-grained geometric and semantic cues needed to distinguish object parts~\cite{huang2023leo, deng20253dllava}. 
Third, grounding mechanisms often rely on a single-granularity query formulation, forcing object-level and part-level targets to share the same query representation and weakening part-aware grounding~\cite{deng20253dllava, huang2025reason3d, zhu2025llava3d}.

To address these challenges, we propose PAR3D, a part-aware 3D-MLLM framework for unified object- and part-level understanding in 3D scenes. Instead of treating parts as merely finer-grained object targets, our framework models them as functional components embedded within objects and situated in scene context.
Concretely, we improve part-aware 3D understanding from three aspects: data, representation, and grounding. First, we construct \textbf{ScenePart}, a synthetic 3D scene dataset that places part-annotated objects into realistic indoor scenes and provides object and part masks, object--part correspondences, and language instructions in scene context. 
Second, we develop \textbf{Part-Aware 3D Representation Learning} on top of a pretrained 3D foundation encoder, adapting the visual backbone to capture fine-grained geometric and semantic features of object parts. 
Third, we introduce \textbf{Hierarchical Segmentation Query Generation}, which generates decoupled segmentation queries for object- and part-level targets to support grounding of parts in relation to their host objects.
Together, these designs form a part-aware 3D-MLLM framework that supports diverse 3D vision-language tasks over both objects and their parts.

In summary, our main contributions are as follows:
\begin{itemize}[leftmargin=10pt]
    \item We construct ScenePart, a synthetic scene-level dataset with object and part masks, object--part correspondences, and language annotations for part-aware 3D-MLLM training and evaluation.
    \item We propose a part-aware 3D-MLLM framework with Part-Aware 3D Representation Learning and Hierarchical Segmentation Query Generation, enabling fine-grained understanding in 3D scenes.
    \item Extensive experiments demonstrate that our framework improves both spatial understanding and grounding at object and part levels across multiple 3D vision-language tasks.
\end{itemize}

\section{Related Works}

\subsection{3D Vision-Language Models for 3D Scenes}

3D vision-language models connect 3D environments with natural language for tasks such as grounding, description, and question answering. Early studies mainly formulate these tasks with dedicated benchmarks and specialist models. For example, ScanRefer~\cite{chen2020scanrefer} and ReferIt3D~\cite{achlioptas2020referit3d} study language-guided object grounding in 3D scenes, Scan2Cap~\cite{chen2021scan2cap} addresses dense caption generation, and ScanQA~\cite{azuma2022scanqa} and SQA3D~\cite{ma2022sqa3d} introduce question answering over 3D environments. 
Beyond these task-specific formulations, subsequent specialist models further improve 3D-language modeling through representation pretraining, stronger cross-modal fusion, and joint modeling across related tasks~\cite{zhu20233dvista, chen2023unit3d, chen2023vote2capdetr, wu20243dstmn}. 
In parallel, recent efforts on semantic-aware neural scene representations, including NeRF- and 3DGS-based methods, also explore how to embed semantic, language, or open-vocabulary information into continuous 3D representations~\cite{qin2023langsplat, qu2024goi, dai2025thgs, lerf2023, qu2023sg, wang2023rip, wang2025look}. 
These methods establish important foundations for 3D language understanding.

With the development of large language models, recent 3D-MLLMs extend language-grounded 3D perception toward instruction following and open-ended reasoning. These methods connect LLMs with 3D scenes through rendered multi-view images, object token representations, or direct point-cloud features~\cite{hong20233dllm,zhu2025llava3d, huang2024chatscene,xu2024pointllm, tang2024minigpt,qi2024gpt4point, huang2025leovl,zheng2025video3dllm}. Recent unified frameworks further support multiple 3D vision-language tasks within a single model~\cite{chen2024ll3da,zhu2024scanreason,deng20253dllava,huang2025reason3d}. Despite their growing generality, these methods still largely rely on object-level supervision and use objects as the primary units for visual representation, language alignment, and grounding. 
As a result, they lack explicit modeling of object--part hierarchy and remain limited in fine-grained part-aware understanding in 3D scenes.

\subsection{3D Part Perception}

3D part perception aims to identify semantically meaningful components of individual 3D objects. Early works mainly study supervised part segmentation on such objects, using part-annotated datasets such as ShapeNetPart~\cite{chang2015shapenet, yi2016shapenetpart} and PartNet~\cite{mo2019partnet}. 
Classic supervised methods~\cite{qi2017pointnet++, li2018pointcnn, phan2018dgcnn, zhao2021pointtransformer, wu2024ptv3} learn point-level representations to predict predefined part categories on single-object point clouds. 
Recent studies improve scalable 3D part segmentation by lifting 2D foundation-model predictions to 3D~\cite{zhou2024pointsam,yang2024sampart3d}, distilling 2D vision-language priors into 3D representations~\cite{umam2024partdistill}, and constructing large-scale 3D part supervision~\cite{ma2025find3d,ma2025p3sam}. These efforts enable promptable, open-set, or text-aligned part decomposition for 3D objects.
Interaction-oriented studies in embodied AI~\cite{xiang2020sapien, mo2021where2act, geng2023gapartnet, wang2025kinematify} further explore object parts for affordance reasoning, articulation modeling, and manipulation. 

Despite these advances, existing approaches mainly focus on part segmentation or decomposition of individual 3D objects, leaving part-aware understanding in complete 3D scenes underexplored. 
In contrast, our work studies object parts in complete 3D scenes, where parts need to be understood in relation to their host objects and scene context. We incorporate part-level supervision into a unified 3D-MLLM for fine-grained reasoning and grounding.

\section{Method}
\label{sec:method}

We present PAR3D, a part-aware unified 3D-MLLM framework for multiple 3D vision-language tasks over objects and parts.
Given a colored point cloud $\mathbf{X} \in \mathbb{R}^{N \times 6}$ and a language instruction, the model generates either a textual response or segmentation masks for referred objects or parts.
Our framework supports part-aware 3D understanding through three components: the proposed dataset \textbf{ScenePart} provides scene-level part supervision, \textbf{Part-Aware 3D Representation Learning} adapts 3D visual backbone to better capture fine-grained features, and \textbf{Hierarchical Segmentation Query Generation} produces granularity-aware grounding tokens for object- and part-level mask prediction.

\subsection{Preliminary: 3D-LLaVA}
\label{sec:preliminary}
Our framework builds on 3D-LLaVA~\cite{deng20253dllava}, a 3D-MLLM that unifies language-based 3D scene understanding and referring segmentation. 
Given a point cloud $\mathbf{X}$, a 3D encoder $\mathcal{E}$ extracts point-level features and aggregates them into superpoint-level features:
\begin{equation}
    \mathbf{F}_e = \operatorname{Pool}(\mathcal{E}(\mathbf{X})),
\end{equation}
where $\operatorname{Pool}(\cdot)$ denotes superpoint pooling, and $\mathbf{F}_e$ consists of $M$ encoder features $\{\mathbf{f}^{e}_i\}_{i=1}^{M}$ over superpoints.
A query decoder $\mathcal{D}$ further refines the superpoint features, using $\mathbf{F}_e$ as both the visual input queries and their key-value features:
\begin{equation}
    \mathbf{F}_d = \mathcal{D}(\mathbf{F}_e, \mathbf{F}_e),
\end{equation}
where $\mathbf{F}_d$ consists of refined features $\{\mathbf{f}^{d}_i\}_{i=1}^{M}$ over the same superpoints.
An MLP projector $\mathcal{P}(\cdot)$ maps the decoder features into the LLM embedding space:
\begin{equation}
    \mathbf{V} = \mathcal{P}(\mathbf{F}_d),
\end{equation}
where $\mathbf{V}$ denotes the 3D visual tokens provided to the LLM.

To enable referring segmentation, 3D-LLaVA uses a special token $[\mathrm{SEG}]$. Its hidden state serves as a mask query and is decoded with the 3D visual features to predict a target mask over superpoints. 
While effective for object-level 3D scene understanding, this framework remains limited in capturing fine-grained part-aware semantics. 
We extend this framework toward part-aware 3D scene understanding by improving its visual representation and segmentation query design.

\begin{figure}[t]
    \centering
    \includegraphics[width=\textwidth]{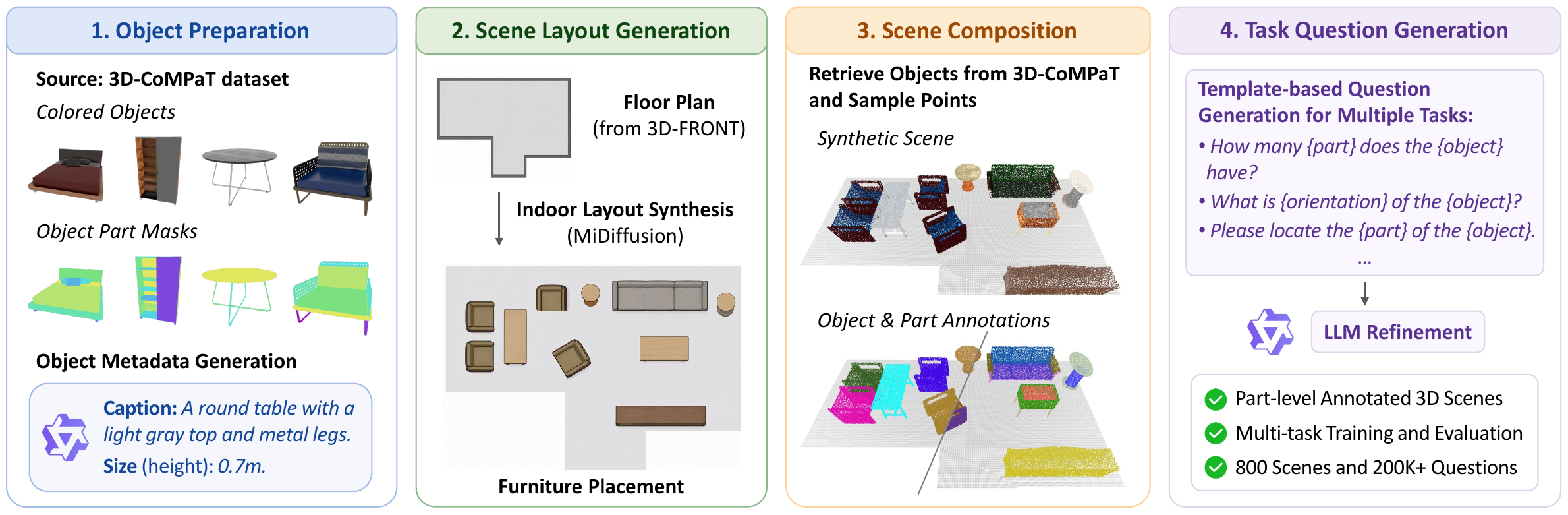}
    \caption{
\textbf{ScenePart Data Construction Pipeline.}
ScenePart composes part-annotated 3D objects into synthesized indoor layouts, producing object- and part-level mask annotations in 3D scenes and multi-task language instructions for training and evaluating part-aware 3D-MLLMs.
    }
    \label{fig:scenepart}
\end{figure}

\subsection{ScenePart Dataset}
\label{sec:method_data}

Scene-level part supervision is largely missing in existing 3D vision-language data. Indoor scene datasets provide realistic object-level annotations~\cite{dai2017scannet, Wald20193rscan, baruch2021arkitscenes} and language tasks~\cite{chen2020scanrefer, azuma2022scanqa, chen2021scan2cap, ma2022sqa3d, zhang2023multi3drefer}, while part-annotated datasets mainly focus on isolated objects \cite{mo2019partnet, liu2023partslip, li20223dcompat, slim20253dcompat++}. To support part-aware 3D-MLLM training, we construct \textbf{ScenePart}, a synthetic 3D scene dataset that places part-annotated objects into indoor scenes and provides object- and part-level supervision in scene context.

Each ScenePart scene is represented as a colored point cloud with object masks, part masks, and object--part correspondences that associate each part instance with its host object. We also provide object descriptions and construct a scene graph that records spatial relationships among objects. In total, ScenePart contains 800 scenes with 21K object masks, 44K part masks, and 273K language-task annotations. Based on these annotations, we build \textbf{ScenePart-200K} as the training set, which contains both visual question answering and referring segmentation instructions. 
For evaluation, we further construct two test splits: \textbf{ScenePart-QA} evaluates part-aware 3D question answering with diverse question types, while \textbf{ScenePart-Seg} evaluates referring segmentation across objects and parts at different granularities. 
In our framework, ScenePart supports both representation adaptation with dense object and part masks, and instruction tuning with language-task annotations.

We construct ScenePart in four steps, as illustrated in Figure~\ref{fig:scenepart}. First, we preprocess part-annotated 3D assets from 3D-CoMPaT~\cite{li20223dcompat, slim20253dcompat++} by filtering object models and normalizing their sizes. We use Qwen3-VL-8B~\cite{bai2025qwen3vl} to estimate object scales and generate object descriptions that are later used for language annotation. 
Second, we generate indoor layouts using MiDiffusion~\cite{hu2026midiffusion} on floor plans from 3D-FRONT~\cite{fu20213dfront}, obtaining furniture placements with category, position, orientation, and scale. 
Third, we instantiate these placements with the preprocessed assets and sample them into point-cloud scenes, where object masks cover the inserted object instances, and part masks are inherited from their internal part annotations.
Finally, we generate language-task annotations from object descriptions, part semantics, object--part correspondences, and scene-graph relations. We use template-based rules for controllability and LLM-based refinement for linguistic diversity.

\begin{figure}[t]
    \centering
    \includegraphics[width=\textwidth]{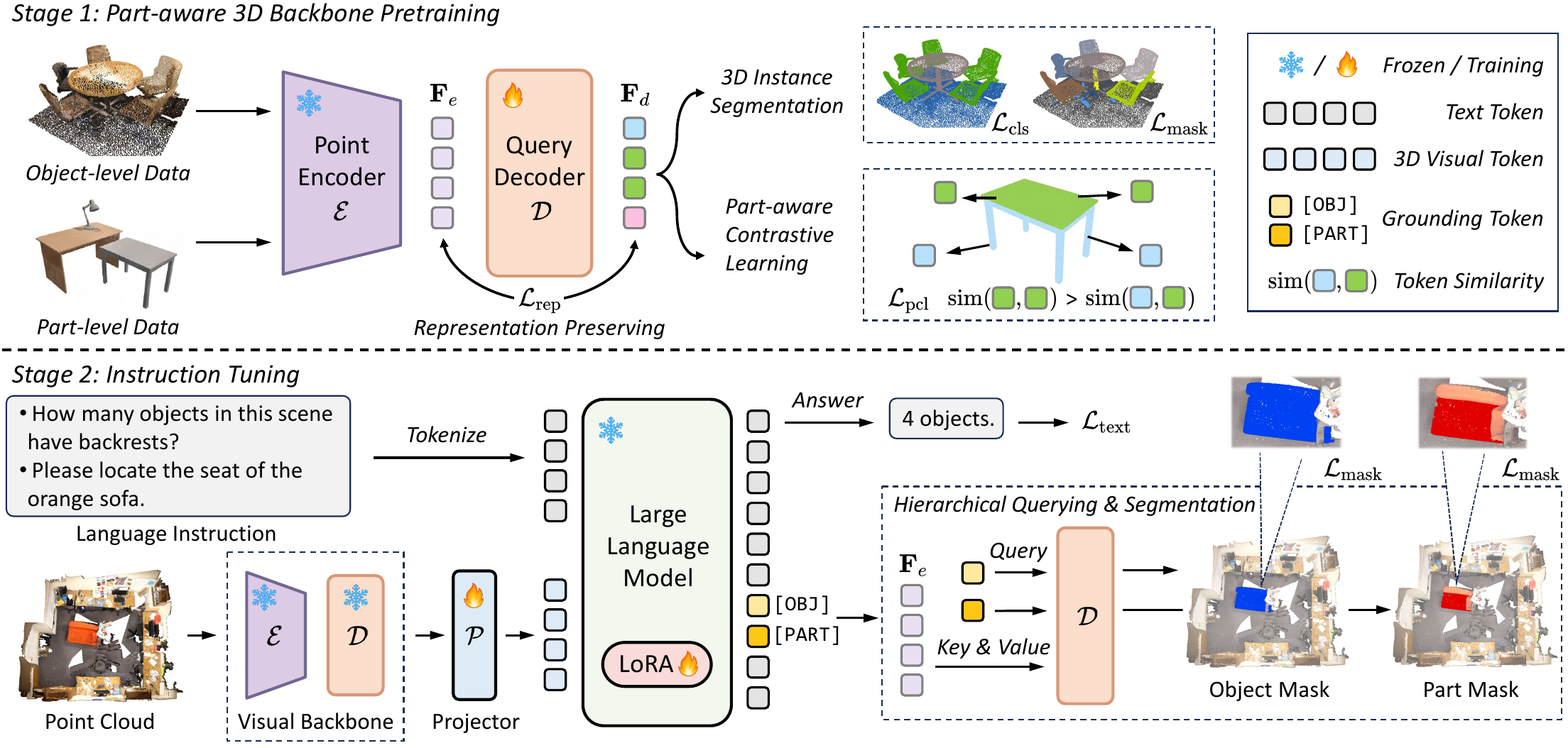}
    \caption{
\textbf{Overall Framework of PAR3D.}
PAR3D is trained with a two-stage scheme.
Stage 1 adapts the 3D visual backbone with object- and part-level supervision through instance segmentation, part-aware contrastive learning, and representation-preserving regularization.
Stage 2 performs instruction tuning on the MLLM using 3D vision-language instruction data.
PAR3D generates textual responses and object or part masks through hierarchical grounding tokens $[\mathrm{OBJ}]$ and $[\mathrm{PART}]$.
    }
    \label{fig:pipeline}
\end{figure}

\subsection{Part-Aware 3D Representation Learning}
\label{sec:method_rep}

Part-aware 3D understanding requires a visual backbone that can capture both scene-level semantics and fine-grained part geometry. Existing 3D-MLLMs commonly rely on 3D encoders or visual backbones adapted from object-level scene understanding, which provide strong object-level perception but offer limited support for fine-grained part representation.
To obtain a more general 3D representation, we instantiate the 3D encoder $\mathcal{E}$ with a pretrained Point Transformer model~\cite{wu2024ptv3, wu2025sonata, zhang2025concerto, zhang2026utonia}. This encoder provides rich geometric and semantic priors, offering a strong foundation for fine-grained recognition in 3D scenes.

In our unified 3D-MLLM, the visual backbone consists of a frozen pretrained encoder $\mathcal{E}$ and a trainable query decoder $\mathcal{D}$. The decoder produces features $\mathbf{F}_d$, which are used for constructing visual tokens. 
A strong encoder alone does not guarantee effective representations for the LLM, as the decoder adapts the features to the model's downstream tasks.
During visual backbone training, we observe that the adapted decoder features can become biased toward mask prediction, potentially deviating from the semantic structure encoded by the frozen encoder. To address this, we introduce two regularization objectives for visual backbone training: Part-aware contrastive learning leverages ScenePart annotations to enhance part-level separability in $\mathbf{F}_d$, while representation-preserving self-distillation aligns the adapted decoder features with the pretrained encoder features.

\paragraph{Part-aware contrastive learning.}

Given a ScenePart scene, we apply contrastive learning to the decoder features $\mathbf{F}_d$ at the superpoint level, compacting features within the same part while separating features from different parts. Let $\mathcal{S}_k$ denote the superpoint indices covered by the $k$-th part mask. For an anchor feature $\mathbf{f}^{d}_i$, $\mathcal{P}(i)$ contains decoder features of other superpoints in the same part mask, and $\mathcal{N}(i)$ contains decoder features of superpoints in different part masks. 
The part-aware contrastive loss follows an InfoNCE objective:
\begin{equation}
    \mathcal{L}_{\mathrm{pcl}}
    =
    - \sum_i
    \log
    \frac{
        \sum_{j \in \mathcal{P}(i)}
        \exp(\mathrm{sim}(\mathbf{f}^{d}_i, \mathbf{f}^{d}_j) / \tau)
    }{
        \sum_{j \in \mathcal{P}(i) \cup \mathcal{N}(i)}
        \exp(\mathrm{sim}(\mathbf{f}^{d}_i, \mathbf{f}^{d}_j) / \tau)
    },
\end{equation}
where $\tau$ is the temperature and $\mathrm{sim}(\cdot,\cdot)$ denotes cosine similarity. This objective improves intra-part feature consistency and strengthens the separability of different part regions.

\paragraph{Representation-preserving self-distillation.}
While part-aware contrastive learning introduces fine-grained supervision, visual backbone adaptation may shift $\mathbf{F}_d$ toward task-specific segmentation representations and weaken the general 3D semantics provided by the pretrained encoder. To control this shift, we use the frozen encoder features $\mathbf{F}_e$ as semantic anchors for the decoder features $\mathbf{F}_d$.

Specifically, for each superpoint, we encourage the decoder feature to remain close to its corresponding encoder feature in the normalized feature space:
\begin{equation}
    \mathcal{L}_{\mathrm{rep}}
    =
    1 -
    \frac{1}{M}
    \sum_{i=1}^{M}
    \mathrm{sim}
    \left(
        \mathbf{f}^{d}_i,
        \mathrm{sg}(\mathbf{f}^{e}_i)
    \right),
\end{equation}
where $M$ is the number of superpoints and $\mathrm{sg}(\cdot)$ denotes the stop-gradient operation. This objective regularizes the visual backbone adaptation by preserving the semantic structure of the pretrained encoder and reducing the task-specific drift of $\mathbf{F}_d$.

\subsection{Hierarchical Segmentation Query Generation}
\label{sec:method_query}

Part-aware grounding requires the language model to generate grounding tokens for targets at different granularities, ranging from whole objects to object parts. 
However, existing 3D-MLLMs with referring segmentation commonly use a single grounding token $[\mathrm{SEG}]$ to represent referred targets of all granularities~\cite{deng20253dllava, huang2025reason3d}.
As a result, object-level and part-level targets are represented by the same type of query, which can cause granularity conflicts and weaken fine-grained part grounding. 
Moreover, part-level grounding requires object-part context, as a part is defined with respect to its host object rather than as an independent region.

We introduce Hierarchical Segmentation Query Generation to reduce this granularity conflict and expose object--part structure in the language-to-grounding interface. 
Instead of using one generic grounding token for all targets, the LLM generates granularity-aware grounding tokens.
For object-level grounding, it generates an object token $[\mathrm{OBJ}]$ to represent the referred object. For part-level grounding, it generates $[\mathrm{OBJ}]$ followed by $[\mathrm{PART}]$, so that host object and target part are represented by separate grounding tokens while remaining coupled in the same language context.

Let $\mathbf{h}_{\mathrm{obj}}$ and $\mathbf{h}_{\mathrm{part}}$ denote the hidden states of $[\mathrm{OBJ}]$ and $[\mathrm{PART}]$, respectively. 
We project these hidden states into segmentation queries with a lightweight MLP $\phi(\cdot)$:
\begin{equation}
    \mathbf{s}_{\mathrm{obj}} = \phi(\mathbf{h}_{\mathrm{obj}}),
    \qquad
    \mathbf{s}_{\mathrm{part}} = \phi(\mathbf{h}_{\mathrm{part}}),
\end{equation}
We reuse the same query decoder $\mathcal{D}$ for mask decoding. 
Specifically, the derived segmentation queries are decoded with the encoder features $\mathbf{F}_e$ as key-value features.
The decoder predicts superpoint-level masks, which are further mapped back to point-level masks:
\begin{equation}
    \hat{\mathbf{m}}_{\mathrm{obj}} = \mathcal{D}(\mathbf{s}_{\mathrm{obj}}, \mathbf{F}_e),
    \qquad
    \hat{\mathbf{m}}_{\mathrm{part}} = \mathcal{D}(\mathbf{s}_{\mathrm{part}}, \mathbf{F}_e).
\end{equation}

During instruction tuning, we supervise the response format according to the target granularity. Object-level referring expressions are trained to generate $[\mathrm{OBJ}]$ and predict the corresponding object mask. Part-level referring expressions are trained to generate both $[\mathrm{OBJ}]$ and $[\mathrm{PART}]$, with mask supervision on the host object and the target part. This joint token and mask supervision encourages the model to preserve object--part structure in its generated grounding tokens and derived segmentation queries while maintaining a unified grounding interface for both object-level and part-level tasks.

\subsection{Training Strategy}
\label{sec:training_strategy}

We train the model in two stages, as illustrated in Figure~\ref{fig:pipeline}. The first stage performs part-aware 3D backbone pretraining, where the query decoder is adapted with object- and part-level supervision while the pretrained point encoder remains frozen. The second stage conducts instruction tuning, where the visual backbone is frozen and the projector and LLM are optimized to support multiple 3D vision-language tasks over objects and their parts.

\paragraph{Stage 1: part-aware 3D backbone pretraining.}
In the first stage, we train the query decoder $\mathcal{D}$ while keeping the pretrained point encoder $\mathcal{E}$ frozen. This stage uses two types of 3D supervision from ScanNet and ScenePart. On ScanNet, we train the query decoder with the 3D instance segmentation objective, which maintains its object-level mask prediction ability:
\begin{equation}
    \mathcal{L}_{\mathrm{inst}}
    =
    \mathcal{L}_{\mathrm{cls}}
    +
    \mathcal{L}_{\mathrm{mask}}.
\end{equation}
Here, $\mathcal{L}_{\mathrm{cls}}$ supervises instance category prediction, and $\mathcal{L}_{\mathrm{mask}}$ supervises the predicted instance masks.
On ScenePart, we use the part-level annotations to apply the part-aware contrastive loss $\mathcal{L}_{\mathrm{pcl}}$ introduced in Section~\ref{sec:method_rep}. In addition, for all training scenes in this stage, we apply the representation-preserving loss $\mathcal{L}_{\mathrm{rep}}$. The overall objective for stage 1 is:
\begin{equation}
    \mathcal{L}_{\mathrm{stage1}}
    =
    \mathcal{L}_{\mathrm{inst}}
    +
    \lambda_{\mathrm{pcl}} \mathcal{L}_{\mathrm{pcl}}
    +
    \lambda_{\mathrm{rep}} \mathcal{L}_{\mathrm{rep}},
\end{equation}
where $\lambda_{\mathrm{pcl}}$ and $\lambda_{\mathrm{rep}}$ balance the auxiliary objectives.

\paragraph{Stage 2: instruction tuning.}
In the second stage, we freeze the 3D visual backbone and train the projector $\mathcal{P}$ and the segmentation MLP $\phi$ together with the LLM using LoRA. We combine existing 3D language understanding datasets with our ScenePart-200K to construct the instruction-tuning data. The existing datasets include ScanRefer~\cite{chen2020scanrefer}, Nr3D~\cite{achlioptas2020referit3d}, Multi3DRefer~\cite{zhang2023multi3drefer}, ScanQA~\cite{azuma2022scanqa}, SQA3D~\cite{ma2022sqa3d}, and Scan2Cap~\cite{chen2021scan2cap}, 
covering object-level referring segmentation, question answering, and captioning. ScenePart further introduces part-aware question answering and referring segmentation instructions.

For language-only tasks, we optimize the standard autoregressive text loss $\mathcal{L}_{\mathrm{text}}$. For grounding tasks, the model is supervised to generate the corresponding grounding tokens and predict masks. Object-level referring expressions generate $[\mathrm{OBJ}]$ and are supervised with object mask annotations, while part-level referring expressions generate both $[\mathrm{OBJ}]$ and $[\mathrm{PART}]$ and are supervised with the host object mask and the target-part mask. We denote the corresponding grounding supervision as $\mathcal{L}_{\mathrm{mask}}$. The stage 2 objective is:
\begin{equation}
    \mathcal{L}_{\mathrm{stage2}}
    =
    \mathcal{L}_{\mathrm{text}}
    +
    \lambda_{\mathrm{mask}} \mathcal{L}_{\mathrm{mask}},
\end{equation}
where the mask loss term is omitted for tasks without grounding supervision.

\section{Experiments}
\label{sec:experiments}

\subsection{Implementation Details}
\label{apx:implement}
The LLM backbone of our work is LLaVA-1.5-7B~\cite{liu2024llava}, following the setting of 3D-LLaVA~\cite{deng20253dllava}. Specifically, the original 3D encoder is replaced with the pretrained Utonia encoder~\cite{zhang2026utonia}. The training proceeds in two stages. In Stage 1, the visual backbone is trained on ScanNet200~\cite{dai2017scannet} and ScenePart for 256 epochs using AdamW with an initial learning rate of $3 \times 10^{-4}$ and polynomial decay. In Stage 2, we perform LoRA-based instruction tuning for 2 epochs on the instruction datasets described in Sec.~\ref{sec:training_strategy}, keeping the main LLM parameters frozen and optimizing the projector MLP and LoRA parameters with AdamW, using an initial learning rate of $2 \times 10^{-4}$ and a cosine annealing schedule. All experiments are conducted on 4 NVIDIA A100 GPUs with 40GB memory.

\subsection{Datasets and Metrics}

\paragraph{Datasets.}
We evaluate our method on the proposed ScenePart test splits and several 3D vision-language benchmarks. For part-aware evaluation, we use ScenePart-Seg and ScenePart-QA introduced in Section~\ref{sec:method_data}. ScenePart-Seg evaluates referring segmentation over objects and parts at different granularities. ScenePart-QA evaluates part-aware 3D question answering with diverse question types involving part semantics, part counting, spatial relations, and object--part associations. To examine object-level performance, we also report results on ScanRefer~\cite{chen2020scanrefer} and Multi3DRefer~\cite{zhang2023multi3drefer} for object-level referring segmentation, ScanQA~\cite{azuma2022scanqa} and SQA3D~\cite{ma2022sqa3d} for 3D question answering, and Scan2Cap~\cite{chen2021scan2cap} for dense caption generation. These benchmarks are based on ScanNet scenes and have been widely adopted in previous works.

\paragraph{Metrics.}
For referring segmentation tasks, we report mean Intersection-over-Union (mIoU) between predicted and ground-truth masks. On ScenePart-Seg, we further report Acc@0.5, i.e., the percentage of predictions with IoU above 50\%, and present results separately for object, coarse-part, and fine-part targets. For open-ended language-generation tasks, including question answering and captioning, we report CIDEr, BLEU-4, METEOR, and ROUGE-L following standard evaluation protocols, abbreviated as C, B-4, M, and R-L in the tables. For SQA3D, which follows a definite-answer setting, we report exact-match accuracy (EM) and refined exact-match accuracy (EM-R) following common evaluation practice.
Higher values indicate better performance for all metrics.

\begin{table}[tbp]
    \caption{
        \textbf{Quantitative Comparison on Object-Level Benchmarks.}
        We compare state-of-the-art methods across 3D referring segmentation, question answering, and dense captioning.
        The best and second-best results are highlighted in \textbf{bold} and \underline{underlined}, respectively.
    }
    \vspace{-0.15cm}
    \renewcommand{\arraystretch}{1.05}
    \centering
    \resizebox{\linewidth}{!}{
    \begin{tabular}{lcccccccccccc}
    \toprule

        \multirow{2}{*}{Methods}&  ScanRefer (val) & \makebox[0.19\linewidth][c]{Multi3DRefer (val)}
        & \multicolumn{4}{c}{ScanQA (val)} & \multicolumn{2}{c}{SQA3D (test)} & \multicolumn{4}{c}{Scan2Cap (val)} \\
        \cmidrule(lr){2-2}\cmidrule(lr){3-3}\cmidrule(lr){4-7}\cmidrule(lr){8-9}\cmidrule(lr){10-13}
        & mIoU$\uparrow$ & mIoU$\uparrow$ & C$\uparrow$ & B-4$\uparrow$ & M$\uparrow$ & R-L$\uparrow$ & EM$\uparrow$ & EM-R$\uparrow$ & 
        \makebox[0.083\linewidth][c]{C@0.5$\uparrow$} & \makebox[0.083\linewidth][c]{B-4@0.5$\uparrow$} & \makebox[0.083\linewidth][c]{M@0.5$\uparrow$} &         \makebox[0.083\linewidth][c]{R-L@0.5$\uparrow$} \\
    
    \midrule
    \multicolumn{13}{l}{  \textbf{\textit{Specialist Models:}}  }  \\
    ScanQA\cite{azuma2022scanqa}   & - & - & 64.9 & 10.1 & 13.1 & 33.3 & 46.6 & - & - & - & - & - \\
    3D-VisTA\cite{zhu20233dvista} & - & - & 69.6 & 10.4 & 13.9 & \textbf{45.7} & 48.5 & - & 61.6 & 34.1 & 26.8 & 55.0 \\
    Scan2Cap\cite{chen2021scan2cap} & - & - & - & - & - & - & 41.0 & - & 39.1 & 23.3 & 22.0 & 44.8 \\
    UniT3D\cite{chen2023unit3d} & - & - & - & - & - & - & - & - & 46.7 & 27.2 & 21.9 & 46.0 \\
    Vote2Cap-DETR~\cite{chen2023vote2capdetr} & - & - & - & - & - & - & - & - & 61.8 & 34.5 & 26.2 & 54.4 \\
    M3DRef-CLIP~\cite{zhang2023multi3drefer} & 35.7 & 32.6& -& - & - & - & - & - & - & - & - & - \\
    3D-STMN~\cite{wu20243dstmn} & 39.5 & - & -& - & - & - & - & - & - & - & - & - \\

    \midrule
    \multicolumn{13}{l}{  \textbf{\textit{Finetuned 3D-MLLMs:}}  }  \\
    3D-LLM \cite{hong20233dllm} & - & - & 69.4 & 12.0 & 14.5 & 35.7 & - & - & - & - & - & - \\
    Scene-LLM \cite{fu2024scenellm} & - & - & 80.0 & 12.0 & 16.8 & 40.0 & 54.2 & - & - & - & - & - \\
    LL3DA \cite{chen2024ll3da} & - & - & 76.8 & 13.5 & 15.9 & 37.3 & - & - & 65.2 & 36.8 & 26.0 & 55.1 \\
    SegPoint \cite{he2024segpoint} & 41.7 & 36.1 & - & - & - & - & - & - & - & - & - & - \\

    \midrule    
    \multicolumn{13}{l}{  \textbf{\textit{Generalist 3D-MLLMs:}}  }  \\
    LEO~\cite{huang2023leo} &- & - & \textcolor{gray}{101.4} & \textcolor{gray}{13.2} & \textcolor{gray}{20.0} & \textcolor{gray}{49.2} & 50.0 & 52.4 & 72.4 & \textbf{38.2} & \underline{27.9} &  \textbf{58.1} \\
    Scene-LLM~\cite{fu2024scenellm} & - & - & 80.0 & 11.7 & 15.8 & 35.9 & 53.6 & - & - & - & - & - \\
    Reason3D~\cite{huang2025reason3d} & 42.0 & - & 73.5 & 12.1 & 15.1 & 37.4 & - & - & - & - & - & - \\
    Chat-Scene~\cite{huang2024chatscene} & - & - & 87.7 & 14.3 & 18.0 & 41.6 & \underline{54.6} & \textbf{57.5} & 77.2 & 36.4 & \textbf{28.0} & \textbf{58.1} \\
    Grounded 3D-LLM~\cite{chen2024grounded3dllm} & - & - & 72.7 & 13.4 & - & - & - & - & 70.6 & 35.5 & - & - \\
    3DGraphLLM~\cite{zemskova20253dgraphllm} & - & - & 88.8 & \underline{15.9} & - & - & \textbf{55.9} & - & \underline{81.0} & 36.5 & - & - \\    
    3D-LLaVA~\cite{deng20253dllava} & \underline{43.3} & \underline{42.7} & \underline{92.6} & \textbf{17.1} & \underline{18.4} & 43.1 & 54.5 & 56.6 & 78.8 & 36.9 & 27.1 & 57.7 \\
    
    \rowcolor{tabblue!10}
    PAR3D (ours) & \textbf{49.9} & \textbf{53.4} & \textbf{95.7} & \underline{15.9} & \textbf{18.9} & \underline{45.0} & \underline{54.6} & \underline{57.3} & \textbf{81.4} & \underline{37.3} & 27.5 & \underline{57.9} \\
    \bottomrule
    \end{tabular}
    }
    \label{tab:object_results}
    \vspace{-0.3cm}
\end{table}

\subsection{Comparison with State-of-the-Art Methods}

We compare PAR3D with existing state-of-the-art approaches across multiple 3D vision-language tasks, covering both object-level and part-level evaluation. 
The compared methods are grouped into three categories.  
\textit{Specialist models} are task-specific methods designed for a particular benchmark or a narrow set of closely related tasks. Most of them do not rely on LLMs.
\textit{Finetuned 3D-MLLMs} are adapted to each dataset or task through task-specific finetuning and often achieve strong in-domain performance. 
\textit{Generalist 3D-MLLMs} are trained once on a diverse collection of 3D vision-language tasks and then evaluated across multiple datasets without task-specific finetuning. 
PAR3D falls into the generalist category and further extends generalist 3D-MLLMs with part-awareness.

\paragraph{Object-level evaluation.}
Table~\ref{tab:object_results} compares different methods on object-level benchmarks, including 3D referring segmentation, question answering, and dense captioning. Overall, PAR3D achieves strong performance across the three groups of compared methods while operating as a generalist 3D-MLLM.
For 3D referring segmentation, PAR3D achieves new state-of-the-art results among 3D-MLLMs on both ScanRefer and Multi3DRefer, outperforming the previous best generalist model, 3D-LLaVA, by absolute gains of 6.6\% and 10.7\%, respectively. 
For language-generation tasks, PAR3D also shows strong performance across question answering and dense captioning. On SQA3D, ScanQA, and Scan2Cap, it achieves competitive performance across most evaluated metrics among all compared methods.
LEO~\cite{huang2023leo}'s ScanQA results are marked in gray and excluded from the main comparison, as they are obtained under a different setting with access to question-related ground-truth objects.
These results demonstrate that the proposed framework improves object-level scene understanding as well, further strengthening the unified multi-task capability of 3D-MLLMs.

\begin{table}[tbp]
    \caption{
\textbf{Quantitative Comparison on the ScenePart Benchmark.}
We compare PAR3D with representative 3D-MLLMs on ScenePart-Seg and ScenePart-QA, covering referring segmentation at different granularities and visual question answering.
The best results are highlighted in \textbf{bold}.
    }
    \vspace{-0.15cm}
    \renewcommand{\arraystretch}{1.05}
    \centering
    \resizebox{\linewidth}{!}{
    \begin{tabular}{lcccccccccccc}
    \toprule

    & \multicolumn{8}{c}{ScenePart-Seg}
    & \multicolumn{4}{c}{ScenePart-QA} \\
    \cmidrule(lr){2-9}\cmidrule(lr){10-13}
    Methods & \multicolumn{2}{c}{Object}
    & \multicolumn{2}{c}{Coarse-Part}
    & \multicolumn{2}{c}{Fine-Part}
    & \multicolumn{2}{c}{All}
    & \multirow{2}{*}{C$\uparrow$}
    & \multirow{2}{*}{B-4$\uparrow$}
    & \multirow{2}{*}{M$\uparrow$}
    & \multirow{2}{*}{R-L$\uparrow$} \\
    \cmidrule(lr){2-3}\cmidrule(lr){4-5}\cmidrule(lr){6-7}\cmidrule(lr){8-9}
    & mIoU$\uparrow$ & Acc@0.5$\uparrow$
    & mIoU$\uparrow$ & Acc@0.5$\uparrow$
    & mIoU$\uparrow$ & Acc@0.5$\uparrow$
    & mIoU$\uparrow$ & Acc@0.5$\uparrow$
    & & & & \\
    \midrule

    3D-LLaVA~\cite{deng20253dllava} & 29.0 & 23.9 & 9.0 & 3.4 & 4.2 & 1.0  &  11.1 & 6.6   & 39.6 & 0.1 & 10.0 & 32.1  \\
    Reason3D~\cite{huang2025reason3d} & 25.5 & 13.8 & 7.4 & 2.3 & 3.9 & 0.8 & 9.6 & 4.0 &  - & - & - & - \\
    3D-LLaVA +ScenePart & 78.4 & 83.3 & 52.8 & 55.7 & 37.6 & 37.4 & 51.8 & 53.9   &  177.2 & 45.6 & 43.7 & 80.6 \\
    
    \rowcolor{tabblue!10}
    PAR3D (ours) & \textbf{89.6} & \textbf{91.5} & \textbf{60.9} & \textbf{66.0} & \textbf{46.0} & \textbf{47.1} & \textbf{60.7} & \textbf{63.5} & \textbf{191.1} & \textbf{61.7} & \textbf{46.9} & \textbf{82.1} \\
    \bottomrule
    \end{tabular}
    }
    \label{tab:part_results}
    \vspace{-0.3cm}
\end{table}

\paragraph{Part-level evaluation.}
Since existing benchmarks do not evaluate part-aware 3D scene understanding, we conduct part-aware evaluation on the proposed ScenePart test datasets, as shown in Table~\ref{tab:part_results}.
We compare PAR3D with representative 3D-MLLMs that support referring segmentation, including 3D-LLaVA and Reason3D.
Reason3D is evaluated only on ScenePart-Seg, as its released model focuses on segmentation and does not support open-ended textual answering for QA tasks.
We also train 3D-LLaVA with ScenePart-200K as a stronger baseline to separate the effect of ScenePart supervision from our model design.
For ScenePart-Seg, we report mIoU and Acc@0.5 at multiple granularities, including object-level targets, coarse parts (\textit{e.g.}, door of a refrigerator), fine-grained parts (\textit{e.g.}, handle of a refrigerator), together with the overall result. 
For ScenePart-QA, we report standard language-generation metrics. Across these settings, PAR3D consistently outperforms representative 3D-MLLMs, demonstrating stronger part-aware 3D scene understanding. Notably, training 3D-LLaVA with ScenePart-200K improves its part-aware performance, indicating that ScenePart provides effective supervision. Nevertheless, PAR3D further improves over this data-enhanced baseline, showing the benefit of the proposed part-aware model design.

\begin{figure}[htbp]
    \centering
    \includegraphics[width=\textwidth]{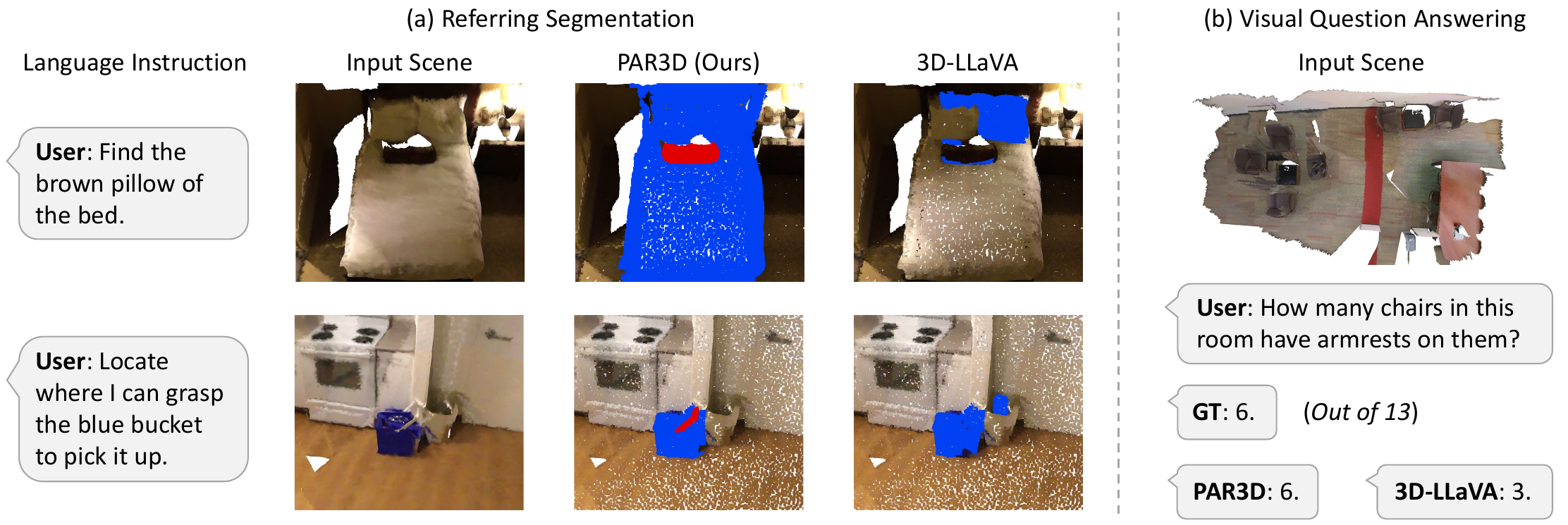}
    \caption{
\textbf{Qualitative Examples of PAR3D.}
We present examples on (a) referring segmentation and (b) visual question answering for part-aware 3D scene understanding.
    }
    \label{fig:qualitative}
\end{figure}

\paragraph{Qualitative results.}
Figure~\ref{fig:qualitative} presents representative examples on real ScanNet scenes with part-aware instructions for referring segmentation and visual question answering. 
Although part-level supervision is only provided by our synthetic ScenePart dataset during training, PAR3D can generalize to real-world 3D scans. 
In segmentation, blue and red masks indicate object- and part-level predictions, respectively. PAR3D predicts fine-grained part masks with host-object context, while 3D-LLaVA often produces coarser object masks or misses the referred part. In part-aware question answering, PAR3D correctly answers a part-aware counting question by recognizing object components such as pillow and bucket handle. 
These examples qualitatively demonstrate PAR3D's ability to handle fine-grained object-part understanding across referring segmentation and question answering. Additional qualitative comparisons are provided in Appendix~\ref{sec:additional_qualitative}.

\subsection{Ablation Studies}

We conduct ablation studies to analyze the contribution of each component in our framework. 
Since our main designs focus on visual representation learning and hierarchical segmentation query generation, we report mIoU on ScenePart-Seg and ScanRefer to evaluate part- and object-level grounding.
As shown in Table~\ref{tab:ablation}, training 3D-LLaVA with ScenePart-200K improves ScenePart-Seg mIoU from 11.1\% to 51.8\%, highlighting the importance of ScenePart supervision for part-aware learning. The improvement on ScanRefer indicates that incorporating ScenePart does not harm object-level grounding.
Beyond data supervision, part-aware 3D representation learning further improves both benchmarks.
The pretrained 3D encoder provides a stronger visual foundation, while the representation-preserving loss and part-aware contrastive loss further adapt the features for fine-grained representations. Finally, hierarchical segmentation query generation achieves the best results on both ScenePart-Seg and ScanRefer, demonstrating that using separate yet coupled segmentation queries for objects and parts benefits fine-grained part grounding as well as object-level grounding.

\begin{table}[htbp]
    \caption{
\textbf{Ablation Studies on Referring Segmentation.}
We evaluate the contribution of each component on ScenePart-Seg and ScanRefer, covering part-aware and object-level referring segmentation.
The best results are highlighted in \textbf{bold}.
    }
    \vspace{-0.15cm}
    \renewcommand{\arraystretch}{1.05}
    \centering
    \resizebox{0.58\linewidth}{!}{
    \begin{tabular}{lcc}
    \toprule
    
        & \makebox[0.15\linewidth][c]{ScenePart-Seg} &  \makebox[0.15\linewidth][c]{ScanRefer} \\
        \multirow{-2}{*}{Methods} & mIoU$\uparrow$  & mIoU$\uparrow$  \\
    \midrule
    3D-LLaVA (baseline)  & 11.1 & 43.3   \\
    + ScenePart Data  & 51.8 &  44.4  \\
    + Pretrained 3D Encoder  & 54.9 &  47.4  \\
    $\quad $ + Representation-Preserving Loss  & 58.7 &  49.1  \\
    $\quad $ + Part-Aware Contrastive Loss  & 59.4 &  49.2  \\
    + Hierarchical Segmentation Query  & \textbf{60.7} &  \textbf{49.9}  \\
        

    \bottomrule
    \end{tabular}
    }
    \label{tab:ablation}
    \vspace{-0.3cm}
\end{table}

\section{Conclusion}
\label{sec:conclusion}

We introduce PAR3D, a unified 3D-MLLM with part-aware representation for fine-grained 3D scene understanding. To extend 3D understanding beyond object level, we construct ScenePart, a synthetic scene-level dataset with object and part masks, object--part correspondences, and language instructions.
Building on ScenePart, we develop part-aware 3D representation learning to improve fine-grained part representations, and introduce hierarchical segmentation query generation to ground part targets via hierarchical queries.
Experiments on ScenePart and standard 3D vision-language benchmarks show that PAR3D substantially improves part-aware question answering and segmentation while maintaining strong object-level performance. We hope this work encourages future unified 3D-MLLMs toward fine-grained perception, reasoning, and grounding over objects and their parts.

\paragraph{Limitations}
Although PAR3D improves part-aware 3D scene understanding, several limitations remain. 
ScenePart provides object- and part-level supervision through synthesized indoor scenes, but may still have a domain gap from real-world 3D scans and is limited by the source object and part categories. Future work may extend PAR3D to real-scene annotations, open-vocabulary part categories, and embodied scenarios requiring more complex object-part reasoning and interaction.

{
\small
\bibliographystyle{abbrvnat}
\bibliography{reference.bib}
}


\newpage
\appendix

\section{Evaluation Metrics}
\label{apx:metrics}

We evaluate PAR3D with existing approaches on referring segmentation, visual question answering, and dense captioning tasks. Since these tasks involve different output formats, we adopt task-specific evaluation metrics following prior 3D scene-language benchmarks.

\paragraph{Referring segmentation metrics.}
For referring segmentation, including ScenePart-Seg, ScanRefer~\cite{chen2020scanrefer}, and Multi3DRefer~\cite{zhang2023multi3drefer}, we evaluate the overlap between the predicted mask and the ground-truth mask using intersection-over-union (IoU). For ScenePart-Seg, we additionally report Acc@0.5, which measures the percentage of samples whose predicted mask achieves an IoU greater than 50\% with the ground-truth mask. 

\paragraph{Question answering metrics.}
For visual question answering, we follow the common evaluation protocols of the benchmarks in the 3D-MLLM literature.
For ScenePart-QA and ScanQA~\cite{azuma2022scanqa}, we report CIDEr~\cite{vedantam2015cider}, METEOR~\cite{banerjee2005meteor}, ROUGE-L~\cite{lin2004rouge}, and BLEU-4~\cite{papineni2002bleu}. CIDEr measures consensus between generated answers and reference answers using weighted n-gram similarity. METEOR evaluates text similarity through word alignments based on exact, stem, synonym, and paraphrase matches. ROUGE-L measures similarity based on the longest common subsequence while BLEU-4 evaluates modified n-gram precision up to 4-grams. For SQA3D~\cite{ma2022sqa3d}, we report EM and EM-R~\cite{huang2023leo}. 
EM measures the percentage of predictions that exactly match the ground-truth answers, while EM-R follows a refined exact-match protocol that performs more flexible answer matching.

\paragraph{Dense captioning metrics.}
For dense captioning on Scan2Cap~\cite{chen2021scan2cap}, we report CIDEr, METEOR, ROUGE-L, and BLEU-4 with the IoU threshold of 0.5, denoted as CIDEr@0.5, METEOR@0.5, ROUGE-L@0.5, and BLEU-4@0.5. 
The suffix @0.5 indicates that the captioning score is counted only when the predicted object region matches the ground-truth object with IoU above 50\%.
These metrics jointly evaluate whether the model can both localize the target object and generate a caption that matches the reference descriptions.

\section{Details on ScenePart Dataset Construction}
\label{apx:data_proc}

ScenePart is constructed to provide scene-level supervision for object parts, which is largely absent from existing 3D scene-language datasets. It integrates part-annotated objects into synthesized indoor scenes and provides object masks, part masks, object--part correspondences, scene graphs, and language-task annotations. The construction of ScenePart proceeds in four steps, following the pipeline illustrated in Fig.~\ref{fig:scenepart}.

First, we preprocess part-annotated 3D assets from 3D-CoMPaT~\cite{li20223dcompat, slim20253dcompat++}. We filter object models according to category compatibility and annotation quality, normalize their geometry, and preserve their part labels and object--part correspondences. We also use Qwen3-VL-8B~\cite{bai2025qwen3vl} to estimate object scales, which are used to match and instantiate assets for the room layouts, and to generate object descriptions, which support downstream language-task annotation.

Second, we generate indoor layouts using MiDiffusion~\cite{hu2026midiffusion}, a diffusion-based scene synthesis model trained on 3D-FRONT~\cite{fu20213dfront}. Given a floor plan, MiDiffusion predicts furniture placements, where each placement specifies the object category, spatial position, orientation, and scale. We use floor plans from 3D-FRONT to obtain diverse scene-level spatial configurations.

Third, we instantiate the generated layouts with the preprocessed 3D assets and sample them into complete point-cloud scenes. For each furniture placement, we retrieve a compatible object model according to its category and geometry. For categories where part-annotated assets are unavailable, we use 3D-FUTURE~\cite{fu20213dfuture} models to preserve scene completeness.
During layout instantiation, we further perform category-aware augmentation by manually inserting and placing additional compatible objects to enrich scene diversity and part-category coverage. 
These objects are selected from the preprocessed asset pool according to room type, object scale and spatial compatibility, and are placed in plausible locations under collision constraints.
Object-level masks are inherited from individual object instances, while part-level masks are preserved only from 3D-CoMPaT models with part annotations and maintained together with their corresponding host object masks. We further construct a scene graph from the spatial relationships among objects, which provides structured scene context for annotation generation.

Finally, we generate language-task annotations from the synthesized scenes. For referring segmentation, we create expressions that refer to either whole objects or object parts and provide the corresponding object or part masks as supervision. For visual question answering, we generate questions based on object descriptions, part semantics, object--part correspondences, and spatial relationships derived from the scene graph. We first use template-based rules to ensure correctness and controllability, and then apply LLM-based refinement to improve linguistic diversity and naturalness.

Overall, this process yields complete scene-level annotations that jointly support object-level and part-level supervision. In addition to object masks and part masks, ScenePart explicitly records the correspondence between each part mask and its host object mask. This paired object--part supervision enables models to learn hierarchical grounding, where a target part is grounded together with the object to which it belongs. These annotations are then used to construct the training set and the held-out evaluation splits described in the following section.

\section{Statistics and Splits of ScenePart}
\label{apx:data_detail}

\paragraph{Dataset scale.}
ScenePart contains 800 synthesized indoor scenes, 21K object masks, 44K part masks, and 273K language-task annotations. These annotations cover both object-level and part-level reasoning and grounding tasks. From the generated annotation pool, we sample 200K annotations for training and construct two held-out evaluation splits with 10K samples each.

\begin{table}[ht]
\centering
\caption{\textbf{Overall Statistics of ScenePart.}}
\label{tab:scenepart_stats}
\vspace{-0.15cm}
\renewcommand{\arraystretch}{1.05}
    \begin{tabular}{l c}
    \toprule
    \makebox[0.3\linewidth][c]{Statistic} & \makebox[0.3\linewidth][c]{Count} \\
    \midrule
    Scenes & 800 \\
    Object masks & 21K \\
    Part masks & 44K \\
    Language-task annotations & 273K \\
    \midrule
    \textbf{\textit{Splits:}} & \\
    ScenePart-200K annotations & 200K \\
    ScenePart-QA samples  & 10K \\
    ScenePart-Seg samples  & 10K \\
    \bottomrule
    \end{tabular}
\end{table}

\begin{table}[ht]
\centering
\caption{\textbf{Breakdown of the ScenePart-Seg and ScenePart-QA Test Splits.}}
\label{tab:scenepart_eval_breakdown}
\vspace{-0.15cm}
\renewcommand{\arraystretch}{1.05}
    \begin{tabular}{cc cc}
    \toprule
    \multicolumn{2}{c}{ScenePart-Seg} & \multicolumn{2}{c}{ScenePart-QA} \\
    \cmidrule(lr){1-2} \cmidrule(lr){3-4}
    
    \makebox[0.15\linewidth][c]{Type} & \makebox[0.15\linewidth][c]{Count} & \makebox[0.15\linewidth][c]{Type} & \makebox[0.15\linewidth][c]{Count} \\
    \midrule
    Object & 2K & Part existence & 3K \\
    Coarse-part & 4K & Part counting & 1.5K \\
    Fine-part & 4K & Part color & 2K \\
     &  & Part spatial relation & 2K \\
     &  & Cross-object & 1.5K \\
    \bottomrule
    \end{tabular}
\end{table}

\paragraph{Train/test split.}
We split ScenePart at the scene level to avoid leakage between training and evaluation. ScenePart-200K is constructed from annotations associated with the training scenes. The held-out scenes are used to build two evaluation splits: ScenePart-QA for part-aware visual question answering and ScenePart-Seg for referring segmentation at multiple granularities.

\paragraph{ScenePart-Seg.}
ScenePart-Seg evaluates language-guided mask prediction at three granularities: object, coarse part, and fine part. Object-level expressions refer to complete object instances. Coarse and fine parts are defined according to the part hierarchy provided by 3D-CoMPaT, where coarse parts correspond to higher-level functional components (\textit{e.g.}, door of a refrigerator) and fine parts correspond to lower-level subparts (\textit{e.g.}, handle of a refrigerator). Each sample consists of a scene, a referring expression, and the ground-truth mask of the target object or part.

\paragraph{ScenePart-QA.}
ScenePart-QA evaluates part-aware 3D visual question answering. Its questions are generated from object names, part labels, object--part correspondences, part colors, local part geometry, and scene-level part statistics. The split covers five question types: part existence, part counting, part color, part spatial relation, and cross-object questions. Together, these question types evaluate whether a model can recognize object parts, reason about missing parts, count repeated components, describe local appearance, infer intra-object spatial relations, and aggregate part information across the scene.

\section{Additional Qualitative Comparisons}
\label{sec:additional_qualitative}
We provide additional qualitative examples to complement the quantitative results in the main paper. The examples are grouped into two figures, covering visual question answering and referring segmentation across multiple evaluation datasets.

\begin{figure}[htbp]
    \centering
    \includegraphics[width=\textwidth]{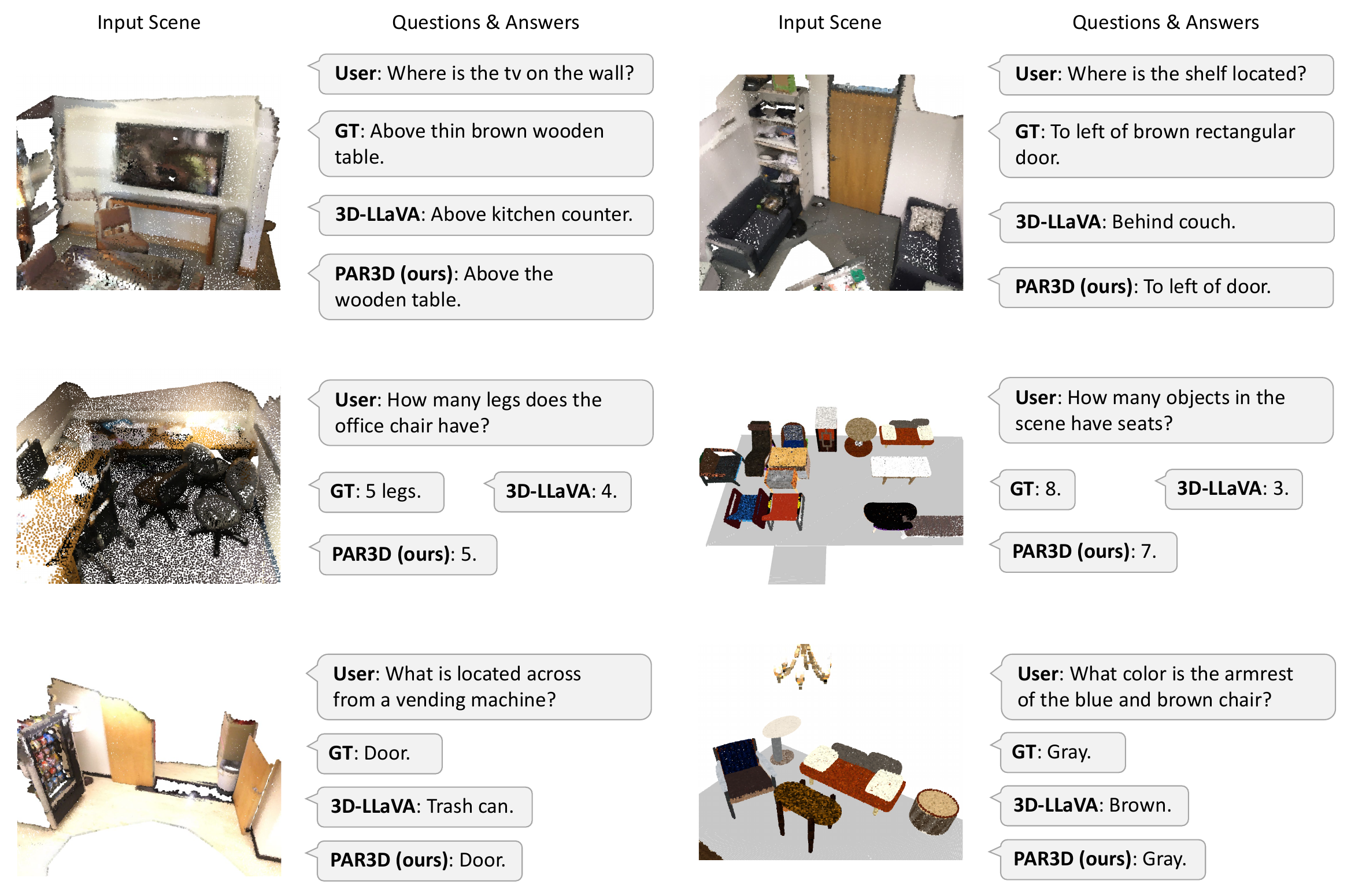}
    \caption{
\textbf{Additional Qualitative Comparisons on Visual Question Answering.}
Each example includes the input scene, the ground-truth answer, the prediction of 3D-LLaVA, and the prediction of PAR3D. PAR3D produces more accurate answers across representative examples from multiple datasets.
    }
    \label{fig:appendix_qa}
\end{figure}

Figure~\ref{fig:appendix_qa} shows additional question answering examples from ScanQA and ScenePart-QA datasets. 
Each example includes the input scene, the ground-truth answer, the answer predicted by PAR3D, and the answer predicted by 3D-LLaVA. 
These cases illustrate that PAR3D can answer questions that require understanding object attributes, object parts, spatial relationships, and scene-level context. Compared to 3D-LLaVA, PAR3D provides more accurate responses on questions requiring fine-grained 3D scene understanding.

\begin{figure}[htbp]
    \centering
    \includegraphics[width=\textwidth]{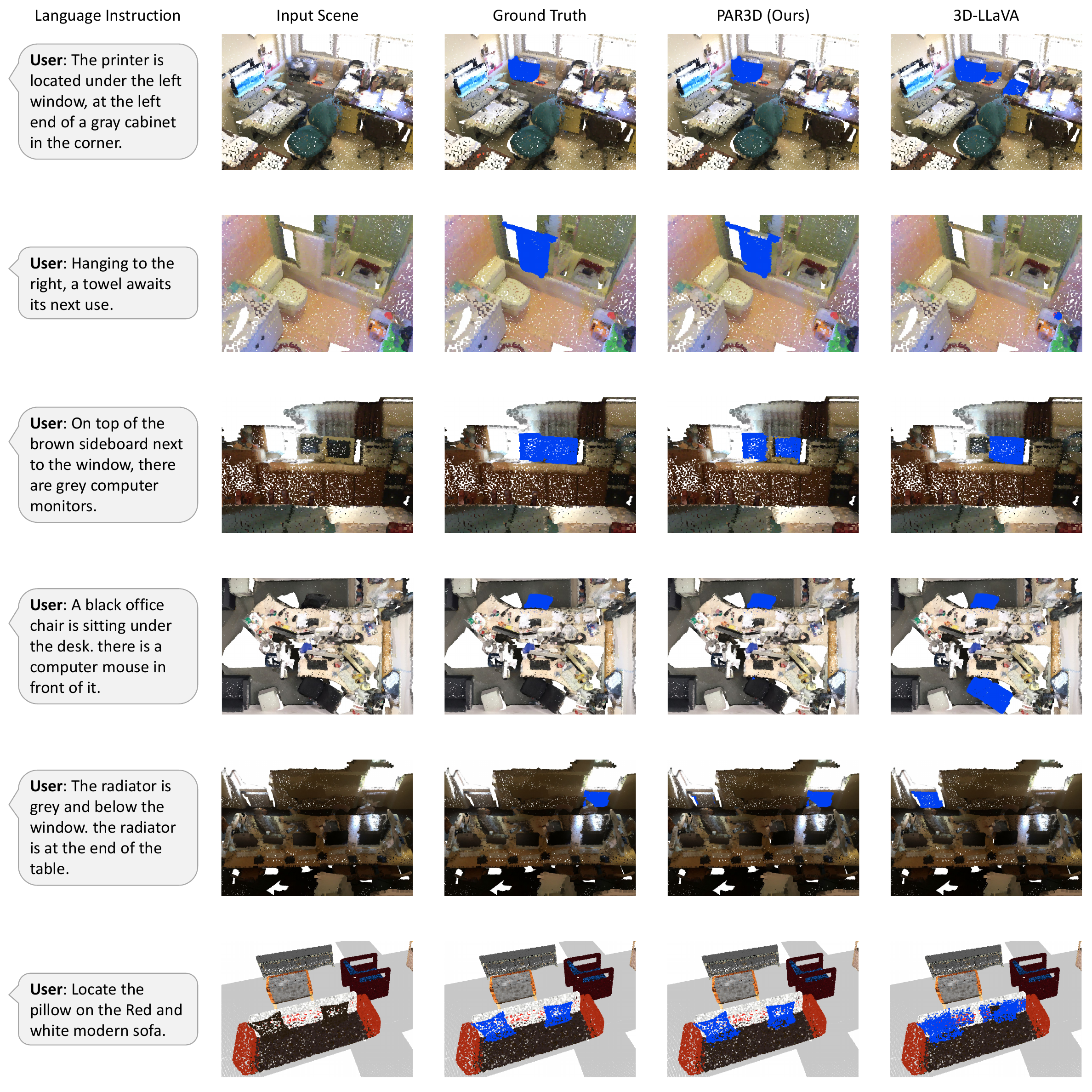}
    \caption{
\textbf{Additional Qualitative Comparisons on Referring Segmentation.}
Each example includes the input scene, the ground-truth mask, the prediction of PAR3D, and the prediction of 3D-LLaVA. 
The target mask is highlighted in blue, regardless of whether the target corresponds to an object or a part.
PAR3D achieves more accurate segmentation across representative examples from multiple datasets.
    }
    \label{fig:appendix_seg}
\end{figure}

Figure~\ref{fig:appendix_seg} shows additional referring segmentation examples from ScanRefer, Multi3DRefer, and ScenePart-Seg datasets. 
Each example includes the input scene, the ground-truth mask, the mask predicted by PAR3D, and the mask predicted by 3D-LLaVA. The results show that PAR3D achieves more accurate segmentation of language-specified targets in 3D scenes across multiple granularities and benchmark settings.

\end{document}